\documentclass{article} 
\usepackage{iclr2021_conference,times}

\usepackage{amsmath,amsfonts,bm}

\def\eqref#1{equation~\ref{#1}}

\def\1{\bm{1}}

\DeclareMathAlphabet{\mathsfit}{\encodingdefault}{\sfdefault}{m}{sl}
\SetMathAlphabet{\mathsfit}{bold}{\encodingdefault}{\sfdefault}{bx}{n}

\usepackage{hyperref}
\usepackage{url}
\usepackage{graphicx}
\usepackage{svg}
\usepackage{tabularx}
\usepackage{booktabs}
\usepackage{placeins}
\usepackage{enumitem}
\usepackage{xcolor}    
\title{WanToFight: Real-Time Generative Game Engine for Multi-Player Combat Interaction}

\author{Li Hu, Guangyuan Wang, Peng Zhang, Bang Zhang\\
Tongyi Lab, Alibaba} 

\iclrfinalcopy 
\makeatletter
\renewcommand{\headrulewidth}{0pt} 
\begin{document}

\maketitle

\begin{abstract}
We present \textbf{WanToFight}, a generative game engine that simulates real-time, two-player \textit{The King of Fighters '97 (KOF~'97)} gameplay from keyboard input. Prior generative game engines target either single-player first-person settings or non-real-time cooperative scenarios; multi-player control, real-time inference, complex physical interaction, and adversarial gameplay have not been jointly addressed. WanToFight closes this gap with three components built on the Wan-1.3B video diffusion transformer: a streaming autoregressive generator with block-causal attention and a rolling KV cache; a visually grounded \emph{Player Association} module that binds each player's keyboard signal to a character identity; and a gated, locally causal keyboard injection module trained with a single-player-to-full-gameplay curriculum. A four-step DMD-distilled student paired with a pruned VAE decoder sustains $30$\,FPS at $512 \times 384$ on a single NVIDIA RTX~5090 over the duration of a complete match. To our knowledge, WanToFight is the first generative game engine to combine multi-player control, real-time inference, complex physical interaction, and adversarial gameplay in one system.
Project Page: {\color{blue}\url{https://humanaigc.github.io/wantofight/}}
\end{abstract}

\section{Introduction}
\label{intro}

Traditional game development relies on three independently authored components: hand-crafted assets, hard-coded game logic, and a rendering engine that composes them into video frames in real time. This pipeline is labor-intensive, and the resulting player experience is bounded by the rules and assets the developers anticipate in advance. Recent advances in video diffusion models~\citep{hunyuanvideo,sora,gao2025seedance,wan2025} have produced high-fidelity, controllable video synthesis, and a growing body of work has begun to shift from text-to-video generation toward interactive \emph{world simulation}~\citep{lingbo-world,genie3,yume15,hunyuanworld,ucm}, in which real-time control signals such as keyboard inputs condition the generated frames. These developments make a single-model alternative to the traditional pipeline---a \emph{generative game engine}~\citep{position} that synthesizes game logic and visuals jointly from player input---technically plausible for the first time.

Most existing works focus on first-person game videos controlled by arrow keys \citep{gamefactory,thematrix,matrixgame2,hunyuangamecraft2,worldmem}. These studies emphasize camera movement and scene consistency, but often lack the ability to generate complex game logic. More recent efforts have demonstrated richer interactive capabilities: GameNGen~\citep{gamengen} simulates \textit{DOOM} with shooting actions, Oasis~\citep{oasis} simulates \textit{Minecraft} with object movement and construction, and Yan~\citep{yan} reproduces collisions during character movement and jumping. A small but rapidly growing line of work explores genuine multi-player settings: Multiverse~\citep{multiverse} implements a dual-perspective racing game where players' actions influence each other's screens, while two concurrent works, Solaris~\citep{solaris} and MultiWorld~\citep{multiworld}, build cooperative multi-agent video world models in \textit{Minecraft} and \textit{It Takes Two}, respectively. However, these efforts remain confined to either non-real-time generation, cooperative or non-combat scenarios, or single-player physical interactions, and none simultaneously support \emph{multi-player control}, \emph{real-time inference}, \emph{complex physical interaction}, and \emph{adversarial gameplay}---the four properties required for a true interactive fighting-game engine. Tab.~\ref{tab:gge_comparison} summarizes a capability comparison of representative methods.

\begin{figure*}[tb]
  \centering
  \includegraphics[width=1.0\textwidth]{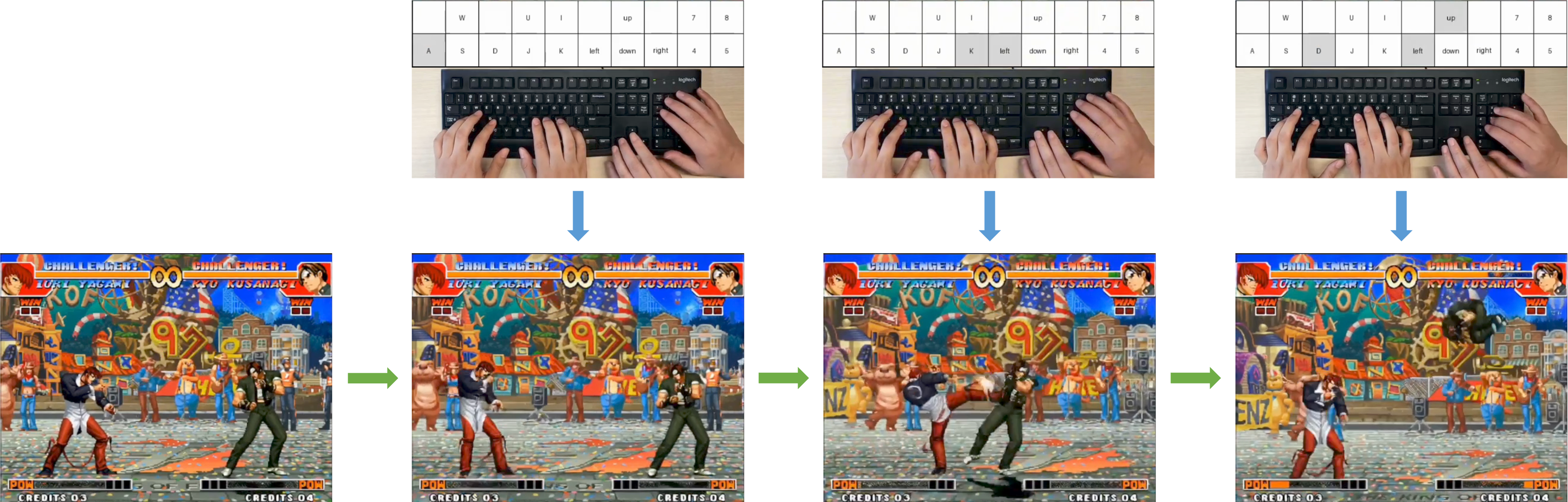}
  \caption{\textbf{WanToFight} generates game visuals based on real-time keyboard inputs, enabling accurate two-player control and complex character interactions.}
  \label{fig:abs}
\end{figure*}

\begin{table*}[t]
\centering
\caption{\textbf{Capability comparison of representative generative game engines and multi-agent video world models.} $\checkmark$ indicates the capability is explicitly supported and demonstrated; \emph{partial} indicates limited or domain-specific support; $\times$ indicates the capability is not supported; ``--'' indicates the column does not apply (e.g., adversarial gameplay is undefined for single-player models). \emph{Real-Time} refers to interactive inference at $\geq 30$\,FPS on a single consumer GPU. \emph{Complex Physics} refers to gameplay in which fast-time-scale physical interactions (e.g., collision, knockback, momentum, vehicle dynamics) materially affect the game outcome, as opposed to incidental physics in voxel or open-world settings. \emph{Adversarial} refers to direct combat between players involving damage and KO mechanics, distinct from competitive racing or cooperative collaboration. To our knowledge, \textbf{WanToFight} is the first generative game engine that simultaneously satisfies all four capabilities.}
\label{tab:gge_comparison}
\resizebox{\textwidth}{!}{%
\begin{tabular}{lccccc}
\toprule
\textbf{Method} & \textbf{Domain} & \textbf{Multi-Player} & \textbf{Real-Time} & \textbf{Complex Physics} & \textbf{Adversarial} \\
\midrule
GameNGen~\citep{gamengen}        & \textit{DOOM}              & $\times$     & $\checkmark$ & partial      & --            \\
Oasis~\citep{oasis}              & \textit{Minecraft}         & $\times$     & $\checkmark$ & partial      & --            \\
WorldMem~\citep{worldmem}        & \textit{Minecraft}         & $\times$     & $\times$     & partial      & --            \\
Multiverse~\citep{multiverse}    & \textit{Gran Turismo~4}    & $\checkmark$ & $\times$     & $\checkmark$ & competitive   \\
Solaris~\citep{solaris}          & \textit{Minecraft}         & $\checkmark$ & $\times$     & partial      & partial       \\
MultiWorld~\citep{multiworld}    & \textit{It Takes Two}      & $\checkmark$ & $\times$     & partial      & $\times$      \\
\midrule
\textbf{WanToFight (Ours)}       & \textit{KOF~'97}           & $\checkmark$ & $\checkmark$ & $\checkmark$ & $\checkmark$ \\
\bottomrule
\end{tabular}%
}
\end{table*}

We present \textbf{WanToFight}, a generative game engine that simulates two-player gameplay over a complete fighting-game session, as illustrated in Fig.~\ref{fig:abs}. We focus on the classic title \textit{The King of Fighters '97 (KOF~'97)}, which exposes three challenges that prior generative game engines do not jointly address:
\begin{itemize}[leftmargin=*, itemsep=2pt, topsep=2pt]
    \item \textbf{Frame-precise physical interaction.} An attack animation must connect at the exact frame when the opponent is in range, with contact triggering the correct knockback, hit-stun, and KO mechanics. Errors of even a few frames produce visibly wrong gameplay.
    \item \textbf{High-frequency, combinatorial inputs.} Skilled play issues rapid sequences of simultaneous direction-and-action key combinations from both players, requiring the model to fuse fast input streams with the current game state at every step.
    \item \textbf{Multi-player identity binding.} The 16-key joint input space must remain consistently routed to the correct character even as the two characters swap sides, occlude each other, or change pose---a regime in which naive control schemes degrade into systematic identity confusion.
\end{itemize}

To address these challenges, WanToFight contributes three components, each tied to a specific failure mode of prior multi-player video world models:
\begin{itemize}[leftmargin=*, itemsep=2pt, topsep=2pt]
    \item \textbf{Streaming autoregressive game engine.} We adapt a pretrained Wan-1.3B video diffusion transformer into a chunk-wise autoregressive generator with block-causal attention and a rolling KV cache. A three-stage training pipeline---bidirectional pretraining, causal adaptation, and DMD-based distillation---combined with a pruned VAE decoder enables sustained $30$\,FPS generation at $512 \times 384$ on a single NVIDIA RTX~5090.
    \item \textbf{Visually grounded Player Association.} To prevent identity crosstalk under fast inputs and frequent position swaps, each player's keyboard signal is bound to that player's character through a reference-image-driven cross-attention block, using the frozen CLIP encoder reused from Wan-I2V. The serial binding generalizes naturally to $N$ players.
    \item \textbf{Gated, locally causal keyboard injection.} The keyboard input is encoded by a 1D convolution and injected into the DiT through cross-attention every three blocks. A data-dependent sigmoid gate filters uninformative inputs and a size-3 temporal window enforces local causality of control. Combined with a single-player-to-full-gameplay curriculum, this design yields stable two-player control under bursty input streams.
\end{itemize}

Together, these components yield WanToFight, the first generative game engine that combines multi-player control, real-time inference, complex physical interaction, and adversarial gameplay in one system (Tab.~\ref{tab:gge_comparison}), sustaining $30$\,FPS two-player \textit{KOF~'97} matches on a single NVIDIA RTX~5090.

\section{Related Work}

\subsection{Generative Game Engines}

Single-player generative game engines simulate one agent's first-person observation conditioned on its action stream. GameNGen~\citep{gamengen} adapts a Stable-Diffusion-style model to render \textit{DOOM} in real time, Oasis~\citep{oasis} extends a similar formulation to \textit{Minecraft}, and Yan~\citep{yan} explores foundational interactive video generation. A subsequent line of work~\citep{gamefactory,matrixgame,matrixgame2,hunyuangamecraft2,worldmem,yume15} further pushes first-person camera control, long-horizon memory, and environmental dynamics. None of these systems address how multiple human controllers should be jointly modeled.

Multi-player video world models form a much smaller line of work. Multiverse~\citep{multiverse} channel-concatenates dual-perspective frames from a \textit{Gran Turismo} racing game. Concurrent work Solaris~\citep{solaris} models cooperative two-player \textit{Minecraft} with an additive learned player-identity embedding injected at each shared self-attention layer, and trains via a four-stage bidirectional-to-causal-to-Self-Forcing pipeline. MultiWorld~\citep{multiworld} treats multi-agent multi-view world modeling more broadly, using rotary positional encoding along the agent dimension (Agent Identity Embedding) for identity disambiguation and a VGGT-based Global State Encoder for cross-view consistency. WanToFight differs in three respects: (i) it grounds player identity in a frozen visual reference rather than a learned positional or additive embedding, which is robust to the position swaps and visual occlusions characteristic of fighting games; (ii) it targets adversarial gameplay with frame-precise combat rather than cooperative or competitive-racing settings; and (iii) it combines streaming autoregression with few-step distillation to reach interactive frame rates on a single consumer GPU.

\subsection{Streaming Autoregressive Video Generation}

Adapting a bidirectional video diffusion model into a streaming, autoregressive generator is essential for any interactive engine. Diffusion Forcing~\citep{diffusionforcing} introduces independent per-frame timesteps so that autoregressive rollout emerges naturally from training. CausVid~\citep{causvid} converts a bidirectional teacher into a few-step causal student through ODE regression followed by distribution-matching distillation. SelfForcing~\citep{selfforcing} closes the train--test gap further by supervising the student on its own generations rather than ground-truth prefixes, mitigating drift accumulation in long autoregressive rollouts. Our Stage~2 training adopts independent per-chunk diffusion timesteps in the spirit of Diffusion Forcing while replacing dense attention with a block-causal mask that exposes each chunk only to itself and the two preceding chunks. At inference, we maintain a rolling KV cache of the two most recent chunks ($16$ frames) cached at the penultimate denoising step, which we find empirically sufficient for stable streaming over the duration of a complete fighting match.

\subsection{Few-Step Distillation for Diffusion Models}

Reducing the number of denoising steps at inference is critical for real-time generation. Distribution Matching Distillation (DMD)~\citep{dmd} trains a few-step student by minimizing a divergence between the student's and the teacher's distributions at each retained timestep, in contrast to the step-wise teacher--student matching used in earlier consistency-based approaches. We apply DMD in full-parameter form to compress our causal teacher to a four-step student. Combined with a structurally pruned VAE decoder fine-tuned on game-domain frames, this distillation enables WanToFight to sustain $30$\,FPS at $512 \times 384$ on a single NVIDIA RTX~5090---a regime that no prior multi-player video world model has demonstrated.

\section{Method}

\subsection{Game Data Collection}

We aim to collect a large-scale corpus of paired \emph{(game video, keyboard input)} sequences for training. To keep the problem tractable, we focus on the set of \emph{fundamental actions}---movement, jumping, crouching, and basic attacks---and exclude advanced combos and character-specific special moves. The pipeline proceeds in three stages.

\textbf{(1) Manual recording.} A human player plays \textit{KOF~'97} while a background utility logs the rendered frames and the keyboard state in synchronized streams. This stage yields several hours of paired data that bootstrap the subsequent automatic pipeline.

\textbf{(2) Keyboard policy training.} On top of the manually collected sequences, we train a small autoregressive model that predicts the next keyboard state from a sliding window of past states. The objective is to produce \emph{diverse and plausible} input sequences for data augmentation, rather than to maximize in-game performance. The resulting policy generates long keyboard trajectories that approximate the distribution of human play while covering a wider range of input patterns than any single human session.

\textbf{(3) Automated playback and recording.} The generated keyboard trajectories are fed back into the actual game through a controller emulator, and a recording program captures the rendered frames in lockstep with the injected keys. The pipeline runs unattended and preserves frame-level alignment between visuals and inputs.

\begin{figure*}[tb]
  \centering
  \includegraphics[width=1.0\textwidth]{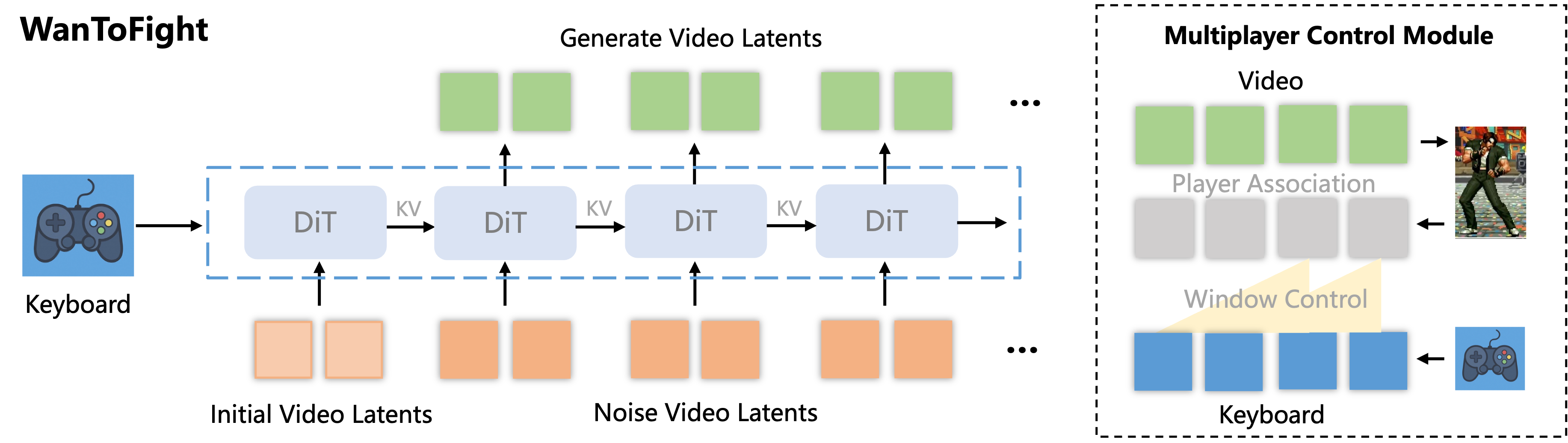}
  \caption{\textbf{Overview of WanToFight.} The model is an autoregressive video diffusion engine that generates gameplay one chunk at a time. Given an initial frame and a stream of keyboard inputs, the model produces the next chunk by conditioning on the previously generated chunk through cached attention states. The \emph{Player Association} module binds each player's keyboard signal to the corresponding character through reference-image-driven attention, and the \emph{Window Attention} mechanism restricts each video latent to attend only to nearby keyboard latents, enforcing local causality of control.}
  \label{fig:game_method1}
\end{figure*}

\subsection{Model Design}

\paragraph{Problem formulation.}
We frame generative game simulation as next-chunk video prediction conditioned on the history of generated frames and player inputs. Let $V = \{x_1, x_2, \dots, x_T\}$ denote a sequence of video frames with $x_t \in \mathbb{R}^{H \times W \times C}$, and let $A = \{a_1, a_2, \dots, a_T\}$ denote the corresponding sequence of keyboard inputs. The model estimates
\begin{equation}
    p_\theta\!\left(x_{t:t+L} \mid x_{<t},\ a_{t:t+L}\right),
\end{equation}
where $L$ is the chunk length. We use $L=8$ frames throughout, corresponding to approximately $0.27$\,s at $30$\,FPS.

\paragraph{Backbone.}
WanToFight builds upon the Wan-1.3B video diffusion transformer~\citep{wan2025,dit,denoising}, which provides strong pretrained visual priors. We adapt this backbone into a streaming, control-conditioned game engine through three components: (i) a chunk-wise autoregressive generator with block-causal attention and KV caching (Sec.~\ref{sec:autoreg}); (ii) a keyboard control module with layer-wise injection, window attention, and gated fusion (Sec.~\ref{sec:control}); and (iii) a Player Association module that grounds each player's input to the corresponding character through a reference image (Sec.~\ref{sec:multi}). The overall architecture is illustrated in Fig.~\ref{fig:game_method1}.

\subsubsection{Streaming Autoregressive Generation}
\label{sec:autoreg}

WanToFight generates gameplay chunk by chunk, where each chunk consists of $L=8$ video frames. At inference time, the model takes an initial game frame as the starting state; from then on, each newly generated chunk becomes the conditioning context for the next, while real-time keyboard inputs are continuously injected. We adapt the bidirectional Wan-1.3B backbone to this streaming setting through three sequential training stages.

\paragraph{Stage 1: Bidirectional pretraining.}
We first fine-tune Wan-1.3B on game data while preserving its original bidirectional attention. Each training sample contains $161$ video frames---roughly $20$ chunks of length $8$---and the keyboard control module described in Sec.~\ref{sec:control} is jointly trained. The relatively long training horizon is intentional: gameplay involves multi-chunk dynamics such as jump arcs, post-hit knockback trajectories, and combo recovery animations, and a model trained on shorter windows tends to lose temporal coherence across these events. This stage equips the backbone with the basic mapping between key presses and character actions while preserving its long-range visual modeling capability.

\paragraph{Stage 2: Causal adaptation.}
We then convert the bidirectional model into a block-causal one to support streaming inference. Following the spirit of Diffusion Forcing~\citep{diffusionforcing}, CausVid~\citep{causvid}, and SelfForcing~\citep{selfforcing}, the dense attention mask is replaced with a \emph{block-causal} mask: tokens within a chunk attend to each other bidirectionally, while tokens across chunks attend only to the current and the two preceding chunks. Different chunks within a sample are assigned independent diffusion timesteps, which prevents the model from collapsing to a single noise level and improves robustness during streaming rollout.

At inference time, generation proceeds one chunk at a time. After producing chunk $C_t$, we cache its keys and values from the penultimate denoising step (e.g., step $3$ of the $4$-step distilled model in Stage~3) and use them as the conditioning context for chunk $C_{t+1}$. We retain the cached keys and values of the two most recent chunks---$16$ frames in total---and discard older history. This rolling window length was determined empirically: it provides sufficient temporal context for smooth chunk-to-chunk transitions while keeping memory bounded regardless of session length.

\paragraph{Stage 3: Efficiency optimization.}
To reach interactive frame rates, we apply two complementary optimizations. First, we distill the causal model with Distribution Matching Distillation~\citep{dmd}, compressing the denoising trajectory to $4$ steps via full-parameter optimization of the student. Second, we replace Wan's native VAE decoder with a pruned variant obtained by structurally reducing the decoder's parameter count and fine-tuning the result on game-domain frames, which reduces per-frame decoding cost without measurably affecting reconstruction quality. Together, these optimizations enable WanToFight to sustain $30$\,FPS at $512 \times 384$ resolution on a single NVIDIA RTX~$5090$.

\subsubsection{Keyboard Control Module}
\label{sec:control}

In two-player mode, WanToFight exposes $16$ keys---four directional (Jump, Left, Crouch, Right) and four action (Light Punch, Light Kick, Heavy Punch, Heavy Kick) per player. The control module specifies how these inputs are encoded, where they enter the DiT, and how their effect on generation is constrained to be temporally local and noise-tolerant.

\paragraph{Input encoding.}
The keyboard state at each video frame is encoded as a $16$-dimensional binary vector indicating which keys are pressed. The resulting per-frame sequence is temporally compressed by a $1$D convolution whose stride matches the temporal compression of the video VAE, so that each keyboard latent aligns with one video latent along the time axis. The compressed sequence is then linearly projected to the cross-attention key/value dimension.

\paragraph{Layer-wise injection schedule.}
Wan-1.3B comprises $30$ transformer blocks. Rather than injecting keyboard features at every block, we insert a cross-attention injection point every three blocks, yielding $10$ injection sites distributed evenly through the network depth. This periodic schedule lets the model integrate keyboard conditioning at multiple representational scales while keeping the per-step compute manageable.

\paragraph{Window attention.}
Game actions are sensitive to the local timing of inputs: an attack frame must respond to the key pressed at \emph{that} latent step, not to keys pressed seconds earlier. We therefore restrict each video latent's attention to a temporal window of size $3$ over the keyboard latents---its own latent step and the two preceding ones. This local-causal masking (i) prevents future keys from leaking into past frames and (ii) discourages the model from averaging over input history that is unrelated to the current frame.

\paragraph{Gated fusion.}
Player input is bursty: long stretches of idleness are punctuated by rapid key sequences, and not every key press has an immediate visual consequence---for example, presses that fall during an unbreakable animation phase. To let the model selectively attenuate weakly informative inputs, we apply a learnable, data-dependent gate to the cross-attention output \emph{after} the keyboard features have attended to the video latents and \emph{before} the result is fused back into the DiT backbone. Concretely, let $h \in \mathbb{R}^{d}$ denote the cross-attention output at each time step; we compute
\begin{equation}
    g = \sigma(W h + b), \qquad \tilde{h} = g \odot h,
\end{equation}
where $\sigma$ is the sigmoid function, $W \in \mathbb{R}^{d \times d}$ and $b \in \mathbb{R}^{d}$ are learned parameters, $\odot$ denotes element-wise multiplication, and $\tilde{h}$ replaces $h$ as the contribution added to the DiT hidden state. Because $h$ already encodes the interaction between keyboard input and the current visual state, the gate can suppress inputs that are irrelevant given the on-screen context (e.g., key presses during an unbreakable animation) while preserving the contribution of decisive inputs.

\begin{figure*}[tb]
  \centering
  \includegraphics[width=1.0\textwidth]{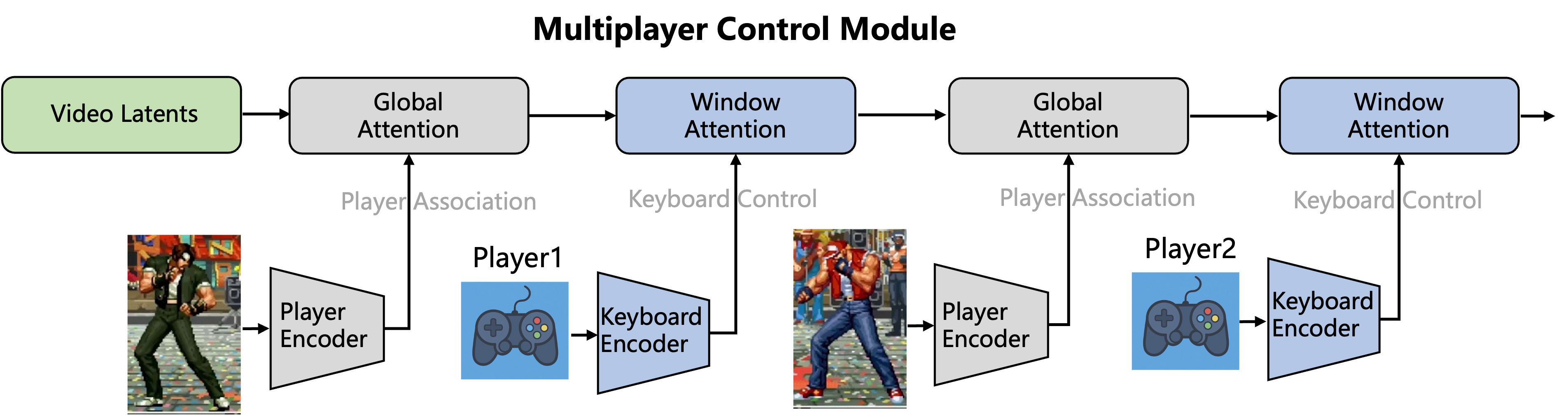}
  \caption{\textbf{Player Association.} For each player, a reference image is encoded by a frozen CLIP encoder to produce an identity embedding. The video latents attend to this embedding through cross-attention (with player feature as keys and values), after which the player's $8$ keys are injected through the keyboard cross-attention pathway. The two players are processed in series within the same forward pass, producing one identity-to-key binding per player.}
  \label{fig:game_method2}
\end{figure*}

\subsubsection{Player Association for Multi-Player Control}
\label{sec:multi}

The $16$ keys of two-player mode pose an identity-binding problem: when the two characters cross sides or visually occlude each other, the model can lose track of which set of $8$ keys controls which character. We refer to this failure mode as \emph{ID crosstalk}, and observe it consistently when all $16$ keys are injected through a single cross-attention pathway. The \emph{Player Association} module resolves crosstalk by grounding each player's input to a reference image of their character.

\paragraph{Reference-image grounding.}
For each player, we provide a pre-set reference image of their character---a clean portrait that captures the character's appearance independent of in-game pose. Each reference is encoded with the frozen CLIP~\citep{clip} vision encoder, yielding a fixed identity embedding per player. Because the references are configured ahead of the session, the encoding cost is incurred once and is amortized across the entire match.

\paragraph{Sequential identity binding.}
For each player, we apply a cross-attention block in which the video latents serve as queries and the player's CLIP feature serves as keys and values, fusing character identity into the video representation. \emph{Immediately after} this binding step, we inject the $8$ keys belonging to that player through the keyboard cross-attention pathway described in Sec.~\ref{sec:control}. The two players are processed in series within the same forward pass: the first player's identity-binding-then-key-injection sub-block precedes the second's, so that each cross-attention round explicitly couples one identity to one key set. This serial design generalizes to any number of players by chaining $N$ identity-binding-and-injection sub-blocks; we show in our experiments that the binding remains stable when characters swap positions or visually overlap.

\subsection{Curriculum Training}

Training directly on full two-player gameplay is unstable: the keyboard signal is sparse relative to the visual complexity, and the model struggles to disentangle character-specific control from background dynamics. We therefore adopt a three-stage curriculum during bidirectional pretraining (Stage~1, Sec.~\ref{sec:autoreg}) that progressively introduces difficulty.

\begin{itemize}[leftmargin=*, itemsep=2pt, topsep=2pt]
    \item \textbf{Single-player actions.} The model first learns the mapping from individual key presses to character animations, using sequences in which only one player is active.
    \item \textbf{Multi-player actions.} The model is then trained on sequences in which both players issue inputs simultaneously, encouraging the Player Association module to establish stable identity-to-key bindings.
    \item \textbf{Full gameplay.} Finally, the model is trained on full-pace match data, learning long-horizon game logic---combos, position swaps, knockback recovery, and the rhythm of attack-and-defense exchanges.
\end{itemize}

This curriculum consistently accelerates convergence and yields more reliable multi-player control under high-frequency inputs than direct training on full gameplay data.

\section{Results}

We evaluate WanToFight on the task of real-time, two-player \textit{KOF~'97} generation, focusing on three properties that follow directly from the challenges identified in Sec.~\ref{intro}: (i) the accuracy of keyboard-driven character control, (ii) the consistency of multi-player identity binding under fast inputs and frequent position swaps, and (iii) the stability of generation over long-duration matches.

\subsection{Experimental Setup}

WanToFight is built on Wan-1.3B and runs at $512 \times 384$ resolution at $30$\,FPS on a single NVIDIA RTX~5090. Generation proceeds chunk by chunk, with each chunk consisting of $8$ frames and corresponding to approximately $0.27$\,s of gameplay. Within a chunk, per-frame keyboard inputs are aligned with the corresponding video latents through the temporal compression described in Sec.~\ref{sec:control}, so multi-key combinations and rapid press sequences are preserved through the compression. Inputs are captured from a standard PC keyboard. In two-player mode, character-movement keys (Jump, Left, Crouch, Right) are mapped to \texttt{WASD} for Player~1 and to the arrow keys for Player~2; the four action keys (Light Punch, Light Kick, Heavy Punch, Heavy Kick) are mapped to \texttt{UIJK} for Player~1 and to \texttt{7}, \texttt{8}, \texttt{4}, \texttt{5} on the numeric keypad for Player~2.

\subsection{Single-Player Keyboard Control}

\begin{figure*}[tb]
  \centering
  \includegraphics[width=1.0\textwidth]{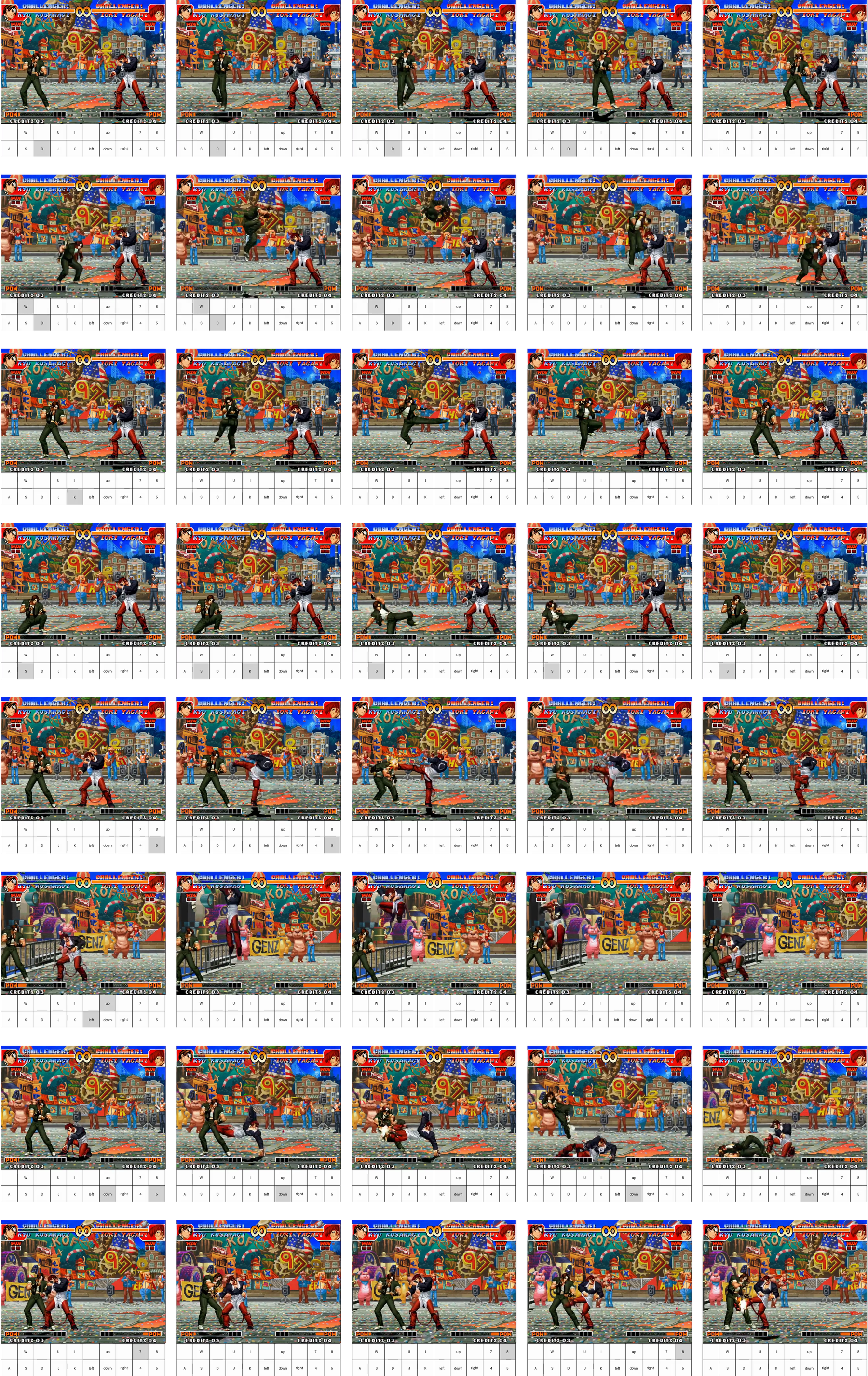}
  \caption{Keyboard-control validation in single-player mode. Each row shows a sequence of generated frames in response to the indicated keyboard input.}
  \label{fig:p12}
\end{figure*}

As shown in Fig.~\ref{fig:p12}, we evaluate WanToFight's response to single-player keyboard input across a range of basic actions, including movement, jumping, crouching, basic attacks, and hitting the opponent. The system responds to player input at chunk-level granularity, and we did not observe input frames being dropped or reordered within a chunk under our default configuration. Composite inputs such as jumping diagonally to the right or attacking while crouching produce the expected on-screen actions, indicating that the keyboard-control module captures combinations of direction and action keys rather than only individual key presses. Character animations remain continuous across chunk boundaries, and rapid input sequences do not produce obvious frame skips or motion artifacts.

\subsection{Multi-Player Identity Binding}

\begin{figure*}[tb]
  \centering
  \includegraphics[width=1.0\textwidth]{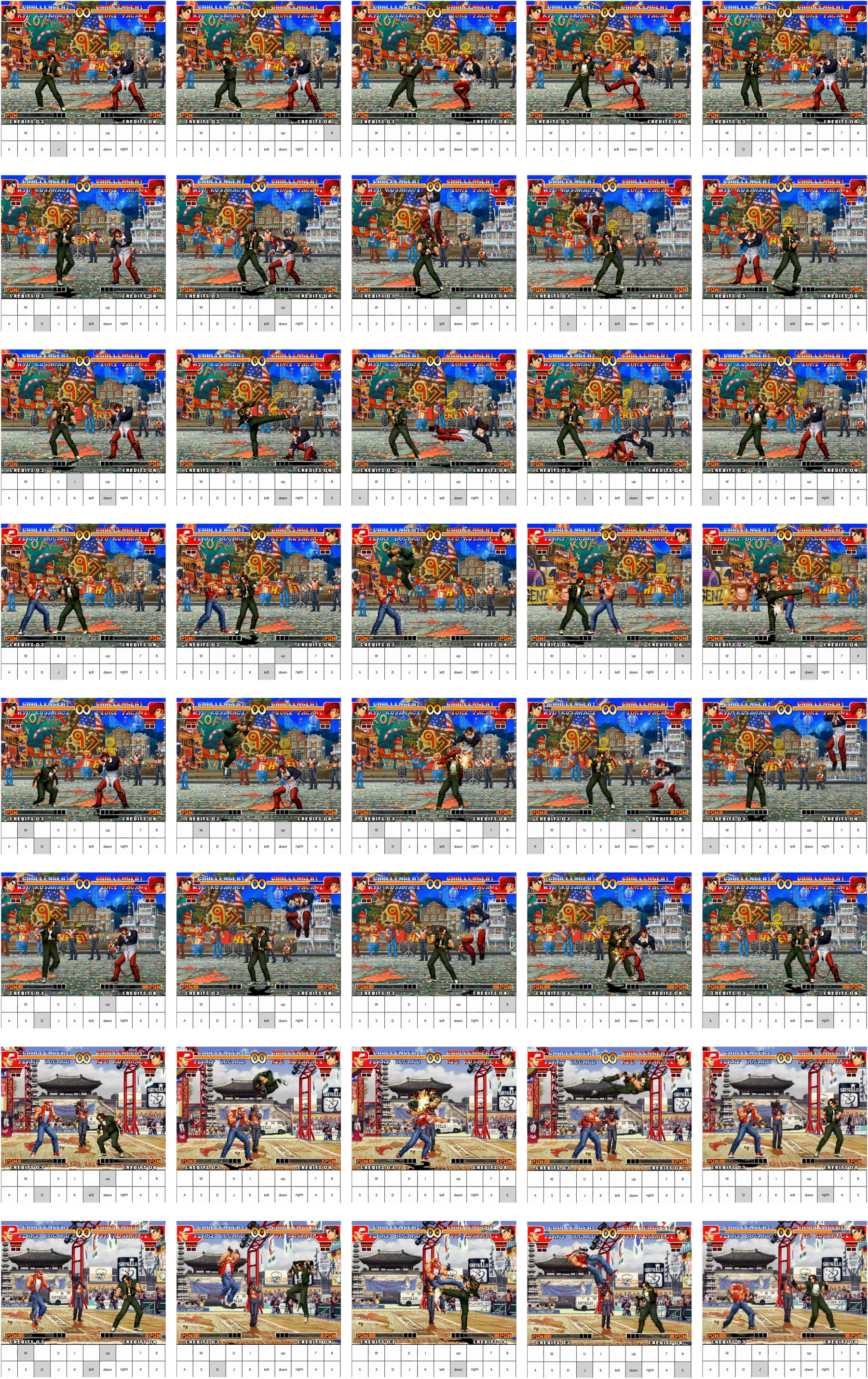}
  \caption{Gameplay validation in two-player mode. The keyboard inputs of the two players consistently drive the correct character even when the two characters cross sides or visually overlap.}
  \label{fig:p34}
\end{figure*}

In two-player mode, the model must consistently route the $16$-key joint input space to the correct character even as the two characters swap sides, occlude each other, or change pose. Fig.~\ref{fig:p34} illustrates this binding under standard gameplay. Each player's inputs drive only their own character throughout the match: when the two characters cross sides on the screen, the keyboard-to-character binding follows the visual identity established at the start of the match rather than the on-screen position. We did not observe identity crosstalk---the failure mode in which, after a position swap, one player's keys begin to drive the wrong character---in any of the matches generated under our default Player Association configuration.

Beyond identity binding, the multi-player generations also exhibit the basic combat physics expected of \textit{KOF~'97}: contact between characters produces appropriate knockback, hit-stun, and downed-state animations, and the relative position of the two characters changes in response to the cumulative effect of sustained pressure. Under high-frequency, simultaneous inputs from both players, the generated motion remains continuous, without obvious frame skips or visual breakdowns.

\subsection{Long-Duration Generation}

By reducing character contact to delay an early KO, we generate continuous gameplay from match start to KO over approximately two minutes without significant visual degradation---substantially exceeding the $161$-frame ($\sim$$5.4$\,s) training horizon and validating the rolling KV cache (Sec.~\ref{sec:autoreg}) for streaming inference well beyond training lengths.

\subsection{Quantitative Evaluation}
\label{sec:quant}

The qualitative observations above are corroborated by two quantitative protocols, each designed to probe one of the core properties identified at the start of this section. Because no prior generative game engine targets the same combination of capabilities (Tab.~\ref{tab:gge_comparison}), we rely on a controlled test set and protocol-internal metrics rather than direct comparison to existing systems.

\paragraph{Action accuracy and design ablation.}
The data-collection pipeline (Stage~3, automated playback) lets us re-render any keyboard trace in the original \textit{KOF~'97} engine, producing a pixel-aligned ground-truth video for that input. We assemble a controlled test set of $500$ input sequences spanning four categories of comparable difficulty: $150$ basic single-player actions, $100$ composite single-player inputs, $100$ two-player sequences without contact, and $150$ two-player sequences with contact. For each sequence we render a ground-truth video in \textit{KOF~'97} and a corresponding generation, and we report perceptual (LPIPS), structural (SSIM), and high-level visual (DINOv2 cosine similarity) similarity averaged over the full test set.

To assess the contribution of each major design choice, we compare WanToFight against a floor baseline that replaces the generated video with the static first frame, an upper-bound multi-step teacher that omits DMD distillation, and three ablations that each remove one design choice at a time (Player Association, gated keyboard injection, or curriculum training). Tab.~\ref{tab:action_acc} reports the results. The $4$-step student matches the multi-step teacher to within $0.02$ LPIPS, $0.03$ SSIM, and $0.03$ DINOv2 similarity, indicating that DMD distillation preserves visual fidelity while delivering the budget required for $30$\,FPS inference. Removing Player Association degrades performance noticeably, primarily on the two-player categories, where keyboard inputs occasionally drive the wrong character; removing the gate makes the model more sensitive to noisy inputs; and skipping the three-stage curriculum yields quality drop, reflecting the difficulty of training directly on full multi-player gameplay.

\begin{table}[t]
\centering
\caption{\textbf{Action accuracy and design ablation.} We compare WanToFight against a floor baseline (repeat first frame), an upper-bound reference (multi-step teacher without DMD distillation), and three ablations that each remove one design choice at a time. Metrics are LPIPS, SSIM, and DINOv2 cosine similarity averaged over $500$ input sequences in our controlled test set. Lower LPIPS and higher SSIM and DINOv2 indicate closer correspondence to the ground-truth \textit{KOF~'97} render driven by the same input.}
\label{tab:action_acc}
\begin{tabular}{lccc}
\toprule
Method & LPIPS $\downarrow$ & SSIM $\uparrow$ & DINOv2 $\uparrow$ \\
\midrule
Repeat first frame (floor)             & $0.49$ & $0.52$ & $0.71$ \\
Multi-step teacher (no DMD, upper bound) & $0.15$ & $0.74$ & $0.85$ \\
\midrule
\textbf{WanToFight (Ours)}             & $\mathbf{0.17}$ & $\mathbf{0.71}$ & $\mathbf{0.82}$ \\
\quad w/o Player Association           & $0.29$ & $0.57$ & $0.69$ \\
\quad w/o Gated keyboard injection     & $0.19$ & $0.69$ & $0.80$ \\
\quad w/o Curriculum training          & $0.22$ & $0.66$ & $0.78$ \\
\bottomrule
\end{tabular}
\end{table}

\paragraph{Multi-player identity-binding consistency.}
We evaluate Player Association by checking, in each generated session, whether the character bound to Player~1 by reference image continues to execute the actions requested by Player~1's keyboard, regardless of position swaps or visual occlusion. To make this checkable from video alone, every test sequence uses an \emph{input-asymmetric} protocol: Player~1 and Player~2 are assigned distinct \textit{KOF~'97} characters and are scripted to issue \emph{visually distinguishable} actions (e.g., one player jumps while the other crouches, or one attacks while the other remains idle). For each generated video, a vision-language judge is then asked a coarse, factual question of the form ``which character is performing $\langle$action$\rangle$: $\langle$P1's character$\rangle$ or $\langle$P2's character$\rangle$?'', and the binding is counted as preserved when the answer matches Player~1's assignment. This reduces the judging task to character recognition plus coarse action recognition, two regimes in which vision-language judges are reliable, and avoids the fine-grained action labelling regime in which they are not.

We construct $500$ test sequences across five scenarios that progressively stress identity binding: $100$ with both characters in static positions, $100$ in which one character jumps over the other, $100$ in which the two characters chase each other across the screen, $100$ with visual overlap or partial occlusion, and $100$ long sessions exceeding $60$\,seconds with periodic input-asymmetric probes throughout. To isolate the contribution of Player Association, we additionally evaluate a baseline that retains the gated keyboard injection module but removes the reference-image identity grounding, concatenating all $16$ keys into a single cross-attention stream so that the model must infer character-to-key correspondence from context alone. We report the percentage of test sequences in which the binding is preserved for each scenario in Tab.~\ref{tab:id_binding}. The baseline performs reasonably only on the static-positions scenario (84\%), where spatial position alone provides a weak identity cue, and drops sharply on every scenario involving a position swap, visual overlap, or long-horizon drift; WanToFight maintains stable binding ($\geq 91\%$) across all four challenging scenarios. The 79--94 percentage-point gap on swap and overlap scenarios directly quantifies the value of grounding identity in a frozen visual reference rather than letting the model learn it implicitly from input concatenation.

\begin{table}[t]
\centering
\caption{\textbf{Multi-player identity-binding consistency, with ablation of Player Association.} Percentage of test sequences in which Player~1's reference-bound character consistently executes Player~1's scripted actions, as judged by a vision-language model under our input-asymmetric protocol (see text). The \emph{w/o Player Association} column is a baseline that concatenates all $16$ keys into a single cross-attention stream with no explicit identity grounding.}
\label{tab:id_binding}
\begin{tabular}{lccc}
\toprule
Scenario & \#~Tests & \textbf{WanToFight} (\%) $\uparrow$ & w/o Player Association (\%) $\uparrow$ \\
\midrule
Static positions               & 100 & $\mathbf{100}$ & $84$ \\
Position swap via jump-over    & 100 & $\mathbf{100}$  & $21$  \\
Position swap via chase/cross  & 100 & $\mathbf{99}$  & $9$  \\
Visual overlap / occlusion     & 100 & $\mathbf{97}$  & $3$  \\
Long session ($>60$\,s)        & 100 & $\mathbf{91}$  & $0$  \\
\midrule
Overall                        & 500 & $\mathbf{97}$  & $23$  \\
\bottomrule
\end{tabular}
\end{table}

\section{Limitations}

WanToFight pushes generative engines into the multi-player, real-time, adversarial regime, but several limitations remain.

\paragraph{Visual fidelity of complex animations.}
\textit{KOF~'97} contains intricate animations such as somersaults, special moves, and complex collision reactions. WanToFight occasionally produces coarse poses or local artefacts in these high-frequency motion regimes. Two factors contribute: the limited visual capacity of the Wan-1.3B backbone, and the modest $512 \times 384$ output resolution, which discards fine-grained sprite detail. Stronger backbones and higher target resolutions would directly address this limitation, at proportionally greater inference cost.

\paragraph{Cumulative drift in long sessions.}
A full \textit{KOF~'97} match can extend beyond two minutes, and a complete training session or series can run for considerably longer. Streaming generation with a two-chunk rolling KV cache is sufficient for the approximately two-minute matches we report, but small errors accumulate across chunks and gradually degrade visual fidelity over longer horizons. 

\paragraph{Single-game scope.}
WanToFight is trained on \textit{KOF~'97} alone. Although the architecture and Player Association mechanism do not depend on \textit{KOF}-specific assumptions, transferring the engine to other fighting titles, or to a single engine that supports many titles, requires both expanded training data and additional work on unified character and action representations across games.

\paragraph{Multi-character scalability beyond two players.}
Our evaluation covers two-player gameplay. The Player Association block extends to $N$ players by chaining $N$ identity-binding-and-injection sub-blocks, but the resulting attention complexity, training-data requirements, and visual capacity for $N \geq 3$ players have not been characterized. Validating these scaling properties is a concrete next step.

\section{Conclusion}

We presented \textbf{WanToFight}, a generative game engine that simulates real-time, two-player \textit{KOF~'97} gameplay from keyboard input. Built on Wan-1.3B, the engine combines streaming autoregressive generation with block-causal attention and a rolling KV cache, a visually grounded \emph{Player Association} module that binds each player's keyboard signal to a character identity through a frozen CLIP encoder, and a gated, locally causal keyboard injection that filters bursty inputs. A four-step DMD-distilled student paired with a pruned VAE decoder sustains $30$\,FPS at $512 \times 384$ on a single NVIDIA RTX~5090 over the duration of a complete match.

WanToFight is the first generative game engine to combine multi-player control, real-time inference, complex physical interaction, and adversarial gameplay in one system (Tab.~\ref{tab:gge_comparison}). The same recipe---reference-image identity grounding paired with streaming, block-causal autoregression and few-step distillation---should extend to other interactive genres requiring real-time per-agent control. Pushing this approach beyond a single title, scaling Player Association beyond two players, and closing the train--test gap with Self-Forcing-style training remain natural directions for future work.


\FloatBarrier

\bibliography{iclr2021_conference}

@String(CVPR= {IEEE Conf. Comput. Vis. Pattern Recog.})

@String(NIPS= {Adv. Neural Inform. Process. Syst.})

@article{wan2025,
      title={Wan: Open and Advanced Large-Scale Video Generative Models}, 
      author={Team Wan and Ang Wang and Baole Ai and Bin Wen and Chaojie Mao and Chen-Wei Xie and Di Chen and Feiwu Yu and Haiming Zhao and Jianxiao Yang and Jianyuan Zeng and Jiayu Wang and Jingfeng Zhang and Jingren Zhou and Jinkai Wang and Jixuan Chen and Kai Zhu and Kang Zhao and Keyu Yan and Lianghua Huang and Mengyang Feng and Ningyi Zhang and Pandeng Li and Pingyu Wu and Ruihang Chu and Ruili Feng and Shiwei Zhang and Siyang Sun and Tao Fang and Tianxing Wang and Tianyi Gui and Tingyu Weng and Tong Shen and Wei Lin and Wei Wang and Wei Wang and Wenmeng Zhou and Wente Wang and Wenting Shen and Wenyuan Yu and Xianzhong Shi and Xiaoming Huang and Xin Xu and Yan Kou and Yangyu Lv and Yifei Li and Yijing Liu and Yiming Wang and Yingya Zhang and Yitong Huang and Yong Li and You Wu and Yu Liu and Yulin Pan and Yun Zheng and Yuntao Hong and Yupeng Shi and Yutong Feng and Zeyinzi Jiang and Zhen Han and Zhi-Fan Wu and Ziyu Liu},
      journal = {arXiv preprint arXiv:2503.20314},
      year={2025}
}

@misc{hunyuanvideo,
      title={HunyuanVideo: A Systematic Framework For Large Video Generative Models}, 
      author={Weijie Kong and Qi Tian and Zijian Zhang and Rox Min and Zuozhuo Dai and Jin Zhou and Jiangfeng Xiong and Xin Li and Bo Wu and Jianwei Zhang and Kathrina Wu and Qin Lin and Junkun Yuan and Yanxin Long and Aladdin Wang and Andong Wang and Changlin Li and Duojun Huang and Fang Yang and Hao Tan and Hongmei Wang and Jacob Song and Jiawang Bai and Jianbing Wu and Jinbao Xue and Joey Wang and Kai Wang and Mengyang Liu and Pengyu Li and Shuai Li and Weiyan Wang and Wenqing Yu and Xinchi Deng and Yang Li and Yi Chen and Yutao Cui and Yuanbo Peng and Zhentao Yu and Zhiyu He and Zhiyong Xu and Zixiang Zhou and Zunnan Xu and Yangyu Tao and Qinglin Lu and Songtao Liu and Daquan Zhou and Hongfa Wang and Yong Yang and Di Wang and Yuhong Liu and Jie Jiang and Caesar Zhong},
      year={2025},
      eprint={2412.03603},
      archivePrefix={arXiv},
      primaryClass={cs.CV},
      url={https://arxiv.org/abs/2412.03603}, 
}

@misc{dmd,
      title={One-step Diffusion with Distribution Matching Distillation}, 
      author={Tianwei Yin and Michaël Gharbi and Richard Zhang and Eli Shechtman and Fredo Durand and William T. Freeman and Taesung Park},
      year={2024},
      eprint={2311.18828},
      archivePrefix={arXiv},
      primaryClass={cs.CV},
      url={https://arxiv.org/abs/2311.18828}, 
}

@inproceedings{clip,
  title={Learning transferable visual models from natural language supervision},
  author={Radford, Alec and Kim, Jong Wook and Hallacy, Chris and Ramesh, Aditya and Goh, Gabriel and Agarwal, Sandhini and Sastry, Girish and Askell, Amanda and Mishkin, Pamela and Clark, Jack and others},
  booktitle={International conference on machine learning},
  pages={8748--8763},
  year={2021},
  organization={PMLR}
}

@article{denoising,
  title={Denoising diffusion probabilistic models},
  author={Ho, Jonathan and Jain, Ajay and Abbeel, Pieter},
  journal={Advances in neural information processing systems},
  volume={33},
  pages={6840--6851},
  year={2020}
}

@article{sora,
  title={Video generation models as world simulators},
  author={Brooks, Tim and Peebles, Bill and Holmes, Connor and DePue, Will and Guo, Yufei and Jing, Li and Schnurr, David and Taylor, Joe and Luhman, Troy and Luhman, Eric and others},
  journal={OpenAI Blog},
  volume={1},
  number={8},
  pages={1},
  year={2024}
}

@inproceedings{dit,
  title={Scalable diffusion models with transformers},
  author={Peebles, William and Xie, Saining},
  booktitle={Proceedings of the IEEE/CVF international conference on computer vision},
  pages={4195--4205},
  year={2023}
}

@misc{genie3,
  title         = {Genie 3: A New Frontier for World Models},
  author        = {Jack Parker-Holder and Shlomi Fruchter},
  year          = {2025},
  url           = {https://deepmind.google/blog/genie-3-a-new-frontier-for-world-models/}
}

@article{gao2025seedance,
  title={Seedance 1.0: Exploring the Boundaries of Video Generation Models},
  author={ByteDance Seed},
  journal={arXiv preprint arXiv:2506.09113},
  year={2025}
}

@article{selfforcing,
  title={Self Forcing: Bridging the Train-Test Gap in Autoregressive Video Diffusion},
  author={Huang, Xun and Li, Zhengqi and He, Guande and Zhou, Mingyuan and Shechtman, Eli},
  journal={arXiv preprint arXiv:2506.08009},
  year={2025}
}

@article{hunyuangamecraft2,
  title={Hunyuan-GameCraft-2: Instruction-following Interactive Game World Model},
  author={Tang, Junshu and Liu, Jiacheng and Li, Jiaqi and Wu, Longhuang and Yang, Haoyu and Zhao, Penghao and Gong, Siruis and Yuan, Xiang and Shao, Shuai and Lu, Qinglin},
  journal={arXiv preprint arXiv:2511.23429},
  year={2025}
}

@article{gamengen,
  title={Diffusion models are real-time game engines},
  author={Valevski, Dani and Leviathan, Yaniv and Arar, Moab and Fruchter, Shlomi},
  journal={arXiv preprint arXiv:2408.14837},
  year={2024}
}

@article{matrixgame,
  title={Matrix-Game: Interactive World Foundation Model},
  author={Yifan Zhang and Chunli Peng and Boyang Wang and Puyi Wang and Qingcheng Zhu and Fei Kang and Biao Jiang and Zedong Gao and Eric Li and Yang Liu and Yahui Zhou},
  journal={arXiv preprint arXiv:2506.18701},
  year={2025}
}

@article{yume15,
  title={Yume-1.5: A Text-Controlled Interactive World Generation Model},
  author={Mao, Xiaofeng and Li, Zhen and Li, Chuanhao and Xu, Xiaojie and Ying, Kaining and He, Tong and Pang, Jiangmiao and Qiao, Yu and Zhang, Kaipeng},
  journal={arXiv preprint arXiv:2512.22096},
  year={2025}
}

@article{matrixgame2,
  title={Matrix-game 2.0: An open-source real-time and streaming interactive world model},
  author={Xianglong He and Chunli Peng and Zexiang Liu and Boyang Wang and Yifan Zhang and Qi Cui and Fei Kang and Biao Jiang and Mengyin An and Yangyang Ren and Baixin Xu and Hao-Xiang Guo and Kaixiong Gong and Size Wu and Wei Li and Xuchen Song and Yang Liu and Yangguang Li and Yahui Zhou},
  journal={arXiv preprint arXiv:2508.13009},
  year={2025}
}

@inproceedings{causvid,
  title={From slow bidirectional to fast autoregressive video diffusion models},
  author={Yin, Tianwei and Zhang, Qiang and Zhang, Richard and Freeman, William T and Durand, Fredo and Shechtman, Eli and Huang, Xun},
  booktitle=CVPR,
  year={2025}
}

@inproceedings{diffusionforcing,
  title={Diffusion forcing: Next-token prediction meets full-sequence diffusion},
  author={Chen, Boyuan and Mart{\'\i} Mons{\'o}, Diego and Du, Yilun and Simchowitz, Max and Tedrake, Russ and Sitzmann, Vincent},
  booktitle=NIPS,
  year={2024}
}

@article{lingbo-world,
  author       = {Zelin Gao and
                  Qiuyu Wang and
                  Yanhong Zeng and
                  Jiapeng Zhu and
                  Ka Leong Cheng and
                  Yixuan Li and
                  Hanlin Wang and
                  Yinghao Xu and
                  Shuailei Ma and
                  Yihang Chen and
                  Jie Liu and
                  Yansong Cheng and
                  Yao Yao and
                  Jiayi Zhu and
                  Yihao Meng and
                  Kecheng Zheng and
                  Qingyan Bai and
                  Jingye Chen and
                  Zehong Shen and
                  Yue Yu and
                  Xing Zhu and
                  Yujun Shen and
                  Hao Ouyang},
  title        = {Advancing Open-source World Models},
  journal      = {CoRR},
  volume       = {abs/2601.20540},
  year         = {2026},
  url          = {https://doi.org/10.48550/arXiv.2601.20540},
  doi          = {10.48550/ARXIV.2601.20540},
  eprinttype    = {arXiv},
  eprint       = {2601.20540},
  bibsource    = {dblp computer science bibliography, https://dblp.org}
}

@article{thematrix,
  author       = {Ruili Feng and
                  Han Zhang and
                  Zhantao Yang and
                  Jie Xiao and
                  Zhilei Shu and
                  Zhiheng Liu and
                  Andy Zheng and
                  Yukun Huang and
                  Yu Liu and
                  Hongyang Zhang},
  title        = {The Matrix: Infinite-Horizon World Generation with Real-Time Moving
                  Control},
  journal      = {CoRR},
  volume       = {abs/2412.03568},
  year         = {2024},
  url          = {https://doi.org/10.48550/arXiv.2412.03568},
  doi          = {10.48550/ARXIV.2412.03568},
  eprinttype    = {arXiv},
  eprint       = {2412.03568},
  bibsource    = {dblp computer science bibliography, https://dblp.org}
}

@article{yan,
  author       = {Deheng Ye and
                  Fangyun Zhou and
                  Jiacheng Lv and
                  Jianqi Ma and
                  Jun Zhang and
                  Junyan Lv and
                  Junyou Li and
                  Minwen Deng and
                  Mingyu Yang and
                  Qiang Fu and
                  Wei Yang and
                  Wenkai Lv and
                  Yangbin Yu and
                  Yewen Wang and
                  Yonghang Guan and
                  Zhihao Hu and
                  Zhongbin Fang and
                  Zhongqian Sun},
  title        = {Yan: Foundational Interactive Video Generation},
  journal      = {CoRR},
  volume       = {abs/2508.08601},
  year         = {2025},
  url          = {https://doi.org/10.48550/arXiv.2508.08601},
  doi          = {10.48550/ARXIV.2508.08601},
  eprinttype    = {arXiv},
  eprint       = {2508.08601},
  bibsource    = {dblp computer science bibliography, https://dblp.org}
}

@article{gamefactory,
  author       = {Jiwen Yu and
                  Yiran Qin and
                  Xintao Wang and
                  Pengfei Wan and
                  Di Zhang and
                  Xihui Liu},
  title        = {GameFactory: Creating New Games with Generative Interactive Videos},
  journal      = {CoRR},
  volume       = {abs/2501.08325},
  year         = {2025},
  url          = {https://doi.org/10.48550/arXiv.2501.08325},
  doi          = {10.48550/ARXIV.2501.08325},
  eprinttype    = {arXiv},
  eprint       = {2501.08325},
  bibsource    = {dblp computer science bibliography, https://dblp.org}
}

@article{worldmem,
  author       = {Zeqi Xiao and
                  Yushi Lan and
                  Yifan Zhou and
                  Wenqi Ouyang and
                  Shuai Yang and
                  Yanhong Zeng and
                  Xingang Pan},
  title        = {{WORLDMEM:} Long-term Consistent World Simulation with Memory},
  journal      = {CoRR},
  volume       = {abs/2504.12369},
  year         = {2025},
  url          = {https://doi.org/10.48550/arXiv.2504.12369},
  doi          = {10.48550/ARXIV.2504.12369},
  eprinttype    = {arXiv},
  eprint       = {2504.12369},
  bibsource    = {dblp computer science bibliography, https://dblp.org}
}

@misc{oasis,
  author    = {Decart and Julian Quevedo and Quinn McIntyre and Spruce Campbell and Xinlei Chen and Robert Wachen},
  title     = {Oasis: A Universe in a Transformer},
  year      = {2024},
  url       = {https://oasis-model.github.io/}
}

@misc{multiverse,
  title={Introducing Multiverse: The First AI Multiplayer World Model},
  author={Enigma team},
  year={2025},
  url={https://enigma.inc/blog}
}

@misc{solaris,
      title={Solaris: Building a Multiplayer Video World Model in Minecraft},
      author={Georgy Savva and Oscar Michel and Daohan Lu and Suppakit Waiwitlikhit and Timothy Meehan and Dhairya Mishra and Srivats Poddar and Jack Lu and Saining Xie},
      year={2026},
      eprint={2602.22208},
      archivePrefix={arXiv},
      primaryClass={cs.CV},
      url={https://arxiv.org/abs/2602.22208},
}

@misc{multiworld,
      title={MultiWorld: Scalable Multi-Agent Multi-View Video World Models},
      author={Haoyu Wu and Jiwen Yu and Yingtian Zou and Xihui Liu},
      year={2026},
      eprint={2604.18564},
      archivePrefix={arXiv},
      primaryClass={cs.CV},
      url={https://arxiv.org/abs/2604.18564},
}

@misc{ucm,
      title={UCM: Unifying Camera Control and Memory with Time-aware Positional Encoding Warping for World Models}, 
      author={Tianxing Xu and Zixuan Wang and Guangyuan Wang and Li Hu and Zhongyi Zhang and Peng Zhang and Bang Zhang and Song-Hai Zhang},
      year={2026},
      eprint={2602.22960},
      archivePrefix={arXiv},
      primaryClass={cs.CV},
      url={https://arxiv.org/abs/2602.22960}, 
}

@misc{position,
      title={Position: Interactive Generative Video as Next-Generation Game Engine}, 
      author={Jiwen Yu and Yiran Qin and Haoxuan Che and Quande Liu and Xintao Wang and Pengfei Wan and Di Zhang and Xihui Liu},
      year={2025},
      eprint={2503.17359},
      archivePrefix={arXiv},
      primaryClass={cs.CV},
      url={https://arxiv.org/abs/2503.17359}, 
}

@misc{hunyuanworld,
      title={HunyuanWorld 1.0: Generating Immersive, Explorable, and Interactive 3D Worlds from Words or Pixels}, 
      author={HunyuanWorld Team and Zhenwei Wang and Yuhao Liu and Junta Wu and Zixiao Gu and Haoyuan Wang and Xuhui Zuo and Tianyu Huang and Wenhuan Li and Sheng Zhang and Yihang Lian and Yulin Tsai and Lifu Wang and Sicong Liu and Puhua Jiang and Xianghui Yang and Dongyuan Guo and Yixuan Tang and Xinyue Mao and Jiaao Yu and Junlin Yu and Jihong Zhang and Meng Chen and Liang Dong and Yiwen Jia and Chao Zhang and Yonghao Tan and Hao Zhang and Zheng Ye and Peng He and Runzhou Wu and Minghui Chen and Zhan Li and Wangchen Qin and Lei Wang and Yifu Sun and Lin Niu and Xiang Yuan and Xiaofeng Yang and Yingping He and Jie Xiao and Yangyu Tao and Jianchen Zhu and Jinbao Xue and Kai Liu and Chongqing Zhao and Xinming Wu and Tian Liu and Peng Chen and Di Wang and Yuhong Liu and Linus and Jie Jiang and Tengfei Wang and Chunchao Guo},
      year={2025},
      eprint={2507.21809},
      archivePrefix={arXiv},
      primaryClass={cs.CV},
      url={https://arxiv.org/abs/2507.21809}, 
}
\bibliographystyle{iclr2021_conference}


\end{document}